\documentclass{article}

\usepackage[preprint]{neurips_2025}

\usepackage[utf8]{inputenc}
\usepackage[T1]{fontenc}
\usepackage{hyperref}
\usepackage{url}
\usepackage{booktabs}
\usepackage{amsfonts}
\usepackage{nicefrac}
\usepackage{microtype}
\usepackage{xcolor}
\usepackage{enumitem}
\usepackage{graphicx}
\usepackage{float}
\usepackage{tabularx}
\usepackage[T1]{fontenc}
\usepackage{underscore}
\usepackage[table]{xcolor}
\usepackage[title]{appendix}

\title{Towards Multi-Turn Dialog Systems for Industrial Asset Operations and Maintenance}

\author{%
  Chengrui Li\thanks{Equal contribution.} \\
  Columbia University \\
  \texttt{cl4750@columbia.edu} \\
  \And
  Rujing Li\footnotemark[1] \\
  Columbia University \\
  \texttt{rl3641@columbia.edu} \\
  \And
  Yitong Bai\footnotemark[1] \\
  Columbia University \\
  \texttt{yb2636@columbia.edu} \\
  \And
  Rui Li\footnotemark[1] \\
  Columbia University \\
  \texttt{rl3586@columbia.edu} \\
}

\begin{document}

\maketitle

\begin{abstract}

Industrial asset operations and maintenance question answering is inherently multi-turn, iterative, and highly dependent on external tool invocation. However, the conventional plan-execute single-agent architecture exhibits clear limitations in maintaining cross-turn context, and reusing intermediate results. In this paper, we present a multi-turn dialog system designed for industrial scenarios based on a supervisor-specialist multi-agent architecture. To alleviate tool invocation bottlenecks, the system incorporates structured artifact reuse, dynamic replanning, and parallel tool execution. Evaluation results show that our system achieves better response quality compared with the baseline, with planning effectiveness increasing by 54.5\% and task completion improving by 37.8\%. System profiling further shows that cross-turn artifact reuse effectively reduces redundant tool invocation, decreasing the tool-time share from 47.3\% to 26.3\% and making turns 2–5 approximately 4.2× faster than the first turn. Code is available at: \url{https://github.com/Coderlicr/Multi-Turn-AssetOps}.

\end{abstract}

% \renewcommand{\thefootnote}{*}
% \footnotetext{Code is available at \url{https://github.com/Coderlicr/Multi-Turn-AssetOps}.}
% \renewcommand{\thefootnote}{\arabic{footnote}}

\section{Introduction}

\subsection{Background and Motivation}

Driven by the emergence of Industry 4.0 technology, continuous monitoring and automated control of physical assets are increasingly employed to improve industrial operations and maintenance (O\&M), thereby facilitating the adoption of predictive maintenance and more energy-efficient processes \cite{rayfield2025react}. However, making effective decisions regarding O\&M tasks frequently involves reasoning across diverse sources, such as databases, equipment knowledge, operation contexts, etc. In real-world diagnostic settings, users rarely pose a single-shot question; rather, they often iterate through their queries by posing additional questions based on prior responses.

This creates a need for multi-turn dialogue systems which can preserve context, reuse intermediary results, and coordinate multiple analytical capabilities. In the field of industry O\&M, tool invocation is often the dominant source of latency and cost. The continuous querying of the same time series data, work order records, alerts, and failure-code mappings not only leads to higher latency but also to instability and inefficiency. Therefore, an effective industrial diagnostic assistant must be able to reuse intermediate artifacts, and adapt its reasoning process as new information becomes available.

\subsection{Problem Statement}

The central problem we focus on is that of developing a robust, efficient and reliable multi-turn dialogue system for industrial O\&M. Existing single-agent plan-execute system \cite{ibm2025assetopsbench} is not well suited for tool-centric industrial diagnosis because it is benchmark-oriented, treating tasks as independent instances without mechanisms for cross-turn memory and artifact reuse.

Specifically, the baseline architecture has three major limitations. Firstly, it does not provide enough planning capabilities necessary for multi-turn, multi-step diagnostic tasks requiring coordination of time-series, work-orders, alerts and maintenance. Secondly, the architecture is unreliable due to frequent tool invocation failures when the system hallucinates tool parameters or fails to recover from the failed executions. Lastly, intermediate results cannot be reused effectively, leading to multiple invocations and higher costs in multi-turn tasks.

To overcome these problems and better meet the demands of practical multi-turn industrial O\&M tasks, this paper explores the shift from single-turn to multi-turn interactions and from the single agent to a multi-agent system that includes a supervisor and multiple specialist agents \cite{shu2024effectivegenaimultiagentcollaboration}. 

\subsection{Objectives and Scope}

Our objective is to design, implement, and assess a multi-turn dialogue system for industrial O\&M that both improves diagnostic quality and reduces unnecessary tool usage. The assessment consists of two parts: evaluation and profiling.

The evaluation focuses on response quality, comparing our system with baseline methods using seven metrics that cover both subjective and objective metrics. Specifically, subjective metrics are assessed by LLM-as-judge \cite{zheng2023judging}, while objective metrics are computed through rule-based validation. 

In addition to response quality, we conduct profiling analysis of our system. The scope is centered on system-level optimization, and we focus on latency and token cost during end-to-end task execution, including the time spent on LLM generation, tool invocation, and different dialogue turns. We also analyze the number of LLM calls and token consumption to understand how multi-agent coordination and artifact reuse affect computational cost and overall system efficiency.

\section{Literature Review}

\subsection{Domain-Specific Language Agents for Industrial O\&M}

Industry 4.0 maintenance leverages real-time sensor data, maintenance records, and domain knowledge \cite{zonta2020predictive}. However, existing works primarily concentrate on the prediction tasks, including fault detection and prognosis. In contrast, the actual end-to-end maintenance workflows remain less explored. ReAct shows that interleaving reasoning steps and environmental actions enhances decision-making while alleviating hallucination for tool-based tasks \cite{yao2023react}. For industrial applications, ReActXen applies this paradigm to natural-language access over SCADA and IoT data, introducing IoTBench for industrial sensor-query evaluation \cite{rayfield2025react}. AssetOpsBench further provides the context of industrial asset operations and maintenance as a benchmark task for agents to reason with perception and control  \cite{ibm2025assetopsbench}. These systems show the growing importance of persistent memory, but they still do not directly address tool-centric industrial scenarios where latency is dominated by repeated tool invocation.

\subsection{Memory and Multi-Turn Interaction}

Industrial multi-turn O\&M differs from single-turn question answering because later user turns often require previous data retrieval, intermediate results, and outputs from tools. To address this problem, the approach of retrieval-augmented generation has been proposed, which combines parameterized priors with non-parameterized external memory \cite{lewis2020retrieval}. Multi-turn O\&M dialogues, however, not only require retrieval of relevant documents but also reusable execution artifacts. MemGPT strives to coordinate the virtual context across memory layers so that agents can maintain extended interactions beyond the context window \cite{packer2023memgpt}. Reflexion records verbal feedback in an episodic memory buffer to support future decision-making \cite{shinn2023reflexion}. Such approaches illustrate the important role of memory in agent behavior; however, they are not designed for tool-centric industrial scenarios, where latency is dominated by repeated tool invocation.

\subsection{Multi-Agent Collaboration}

Complex O\&M tasks need coordinated efforts of various capabilities, including time-series analysis, alert interpretation, data retrieval, and maintenance suggestion. It limits the scalability of single-agent architectures and motivates multi-agent decomposition. General LLM multi-agent frameworks have explored several coordination patterns \cite{wu2023autogen}. Recently, enterprise-oriented work has studied supervisor-specialist hierarchies, where a supervisor plans, delegates, and aggregates outputs from specialized agents \cite{shu2024effectivegenaimultiagentcollaboration}. These findings have encouraged the integration of multi-agent collaboration within end-to-end industrial O\&M discussed in this paper.

\begin{figure*}[!t]
    \centering
    \includegraphics[width=\textwidth]{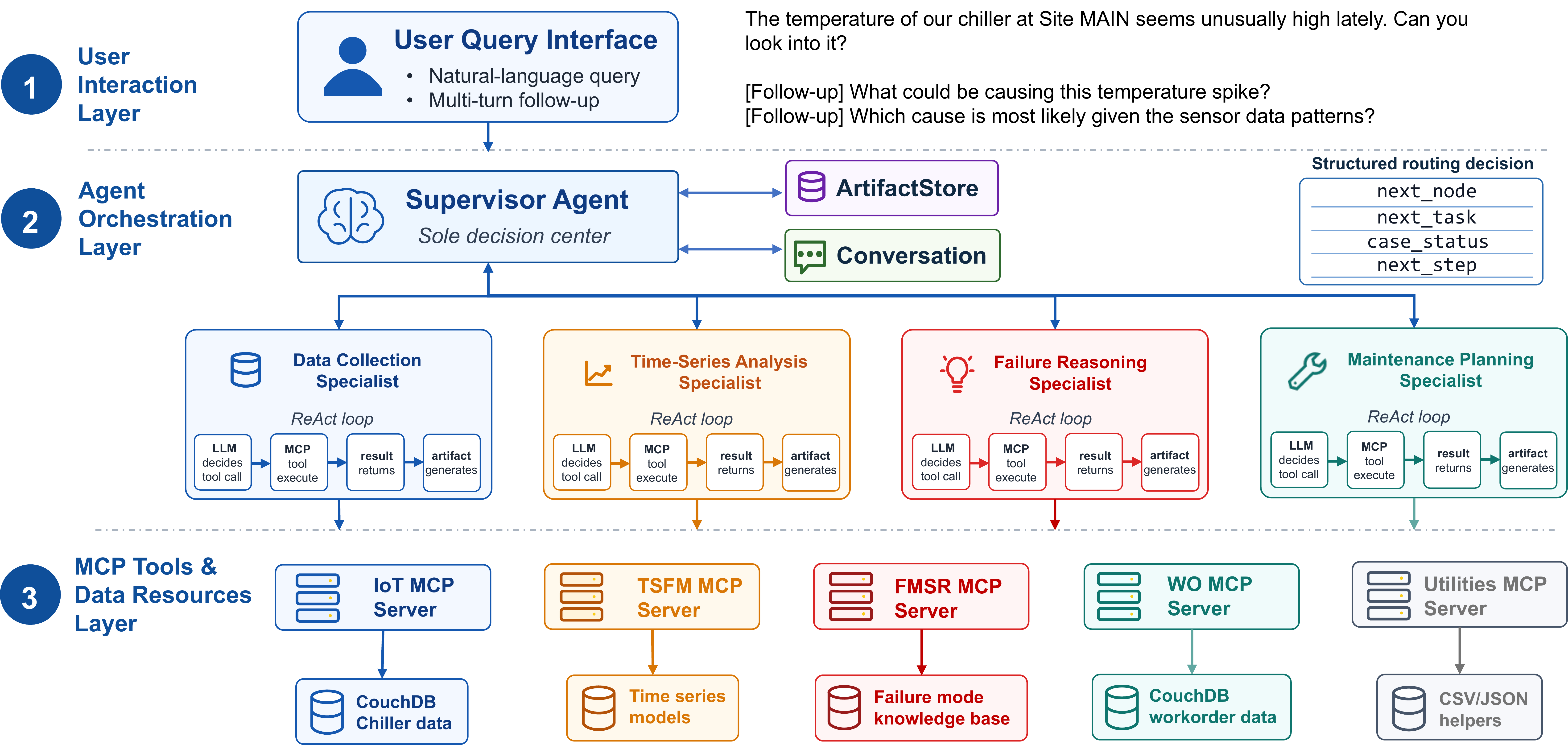}
    \caption{Overview of our supervisor-specialist multi-agent system for industrial asset operations. The system contains three layers: user interaction layer, agent orchestration layer and MCP tools \& data resources layer. Users can input natural language queries, and ask follow-up questions.}
    \label{fig:ss-framework}
\end{figure*}

\section{Methodology}

\subsection{System Overview}

The proposed system adopts a supervisor-specialist multi-agent framework for multi-turn industrial O\&M question answering. As shown in Fig.~\ref{fig:ss-framework}, there are three layers to this multi-agent framework: user interaction layer, agent orchestration layer, and MCP tools and data resources layer. The user interaction layer receives the initial question and any follow-up questions, while maintaining dialog state and reusable artifacts from previous turns. At the agent orchestration layer, there is a supervisor agent and several specialist agents. The supervisor interprets the intention of the user, figures out what evidence has been acquired, breaks down the task into subtasks, and then assigns them to the corresponding specialist. The tools and data resources layer is where external capabilities are made accessible via MCP servers such as IoT sensor retrieval, time-series forecasting, anomaly detection, work-order lookup, event and alert retrieval, and failure-code mapping.

In our implementation, the data collection specialist retrieves asset-level operational evidence, the time-series specialist performs forecasting and anomaly scoring, the failure-reasoning specialist links abnormal observations with possible failure modes, and the maintenance-planning specialist converts the diagnostic evidence into recommended actions. This decomposition follows the supervisor-specialist pattern discussed in prior multi-agent work, where a central coordinator delegates specialized subtasks and aggregates their outputs \cite{shu2024effectivegenaimultiagentcollaboration}.

Each specialist returns a structured artifact rather than only a free-form text answer. An artifact records the asset identifier, time range, invoked tools, key observations, intermediate results, and any assumptions or confidence signals. The supervisor then synthesizes these artifacts into the final response for the current turn and stores them for later turns. This design is particularly important for industrial O\&M dialogs because follow-up questions often refer implicitly to prior evidence, such as ``the same chiller,'' ``that anomaly,'' or ``the previous maintenance record.'' By grounding later reasoning in stored artifacts, the system can preserve context without repeatedly retrieving the same time-series or work-order data.

\subsection{Optimization Techniques}

To address the limitations of the plan-execute baseline, we introduce three system-level optimization techniques. These techniques target the main bottlenecks observed in tool-centric industrial scenarios: repeated external tool calls, fragile tool-argument construction, and inefficient sequential execution.

\textbf{Cross-Turn Artifact Reuse. }Since industrial diagnostic conversations often revisit the same asset, time window, or maintenance evidence, many tool calls in a naive system are redundant. Our system stores the structured outputs generated by specialists. Before invoking a new tool, the supervisor first checks whether an existing artifact already satisfies the current request. If a follow-up question can be answered using previously collected evidence, the supervisor reuses the corresponding artifact directly; if not, it invokes tools only to obtain the information that is absent from existing artifacts.

\textbf{Dynamic Specialist Routing and Replanning. }In the baseline, a single agent generates a plan and follows it in a mostly linear manner. This strategy is fragile when the initial plan omits necessary evidence, selects inappropriate tools, or incorrectly infers dependencies. In contrast, our supervisor re-evaluates the workflow at each turn and after each specialist result. Routing decisions depend on the current user intent, available artifacts, missing evidence, and tool execution status.

% \subsubsection{Parallel Tool Execution}

\textbf{Parallel Tool Execution. }Some diagnostic subtasks require multiple independent data sources, such as retrieving sensor history, recent alerts, historical work orders, and failure-code mappings for the same asset. When these calls do not depend on each other, executing them sequentially increases wall-clock latency unnecessarily. In the advanced supervisor-specialist architecture, independent MCP tool calls within a specialist can be launched concurrently and then normalized into a single structured artifact after all results return.

\subsection{Evaluation Pipeline}

The evaluation module is intended to assess whether our system can not only generate high-quality responses, but also plan effectively, call the right tools, execute successfully, and recover from failures. For each scenario, we consider two kinds of input to the pipeline: the ground truth, and the rollout log files generated by our system and the baseline. The ground truth specifies the expected task intent, associated assets, target evidence, and execution trajectory, while the rollout log files capture the user input, system responses, tool calls and their arguments, as well as their execution status.

The pipeline begins by normalizing the rollout file into a standardized dialog format, a necessary step given the architectural variability of rollout files. The evaluation then proceeds through three branches that operate concurrently. The first branch involves a subjective evaluation that scores the effectiveness of planning, the quality of tool utilization, and whether tasks are completed. The second branch assesses tool-call validity based on schema matching and log parsing. In this case, the branch ensures that each tool call uses a valid tool name, that its arguments conform to the expected schema and types, and that the process completes successfully. The third branch evaluates the recovery behavior by testing the ability to continue towards a final answer despite any retries, or replans.

We use seven metrics to characterize both response quality and execution reliability, as detailed in Appendix~\ref{app:evaluation_metrics}.  
The first three metrics are subjective dialog-level scores in $[0,1]$ and are macro-averaged across dialogs, while the next three metrics are automatic call-level scores and are micro-averaged over tool calls. The last metric measures whether the system can recover from intermediate failures such as invalid arguments, failed tool callings, or inadequate evidence.

\subsection{Profiling Pipeline}

The profiling module evaluates system-level efficiency by comparing three architectures: the plan-execute baseline, the supervisor-specialist architecture, and a parallelized variant of the supervisor-specialist architecture in which specialist agents can execute tools concurrently.
Each of the three architectures undergoes profiling
using the exact same set of 16 multi-turn industrial dialogs.

Profiling is done in three tiers. The first tier captures all calls made to the LLM API, including the model name, prompt tokens, completion tokens, and latency for each completion. The second tier captures all MCP tool invocations within the agents, recording the server name, tool name, latency, and status of each call. The third tier captures database queries within the IoT, and work-order MCP servers, recording the query type, latency, number of documents returned, and any errors.

The following are recorded as dialog-level metrics and shown on the WandB dashboard : wall time, a three-way latency decomposition involving LLM time, tool time, and routing time (the residual after subtracting the first two from wall time), tokens consumed, number of LLM API calls, and per server tool latency. Turn-level metrics track turn duration, turn success, and length of the turn output with a unique turn index across all dialogs to observe how the per-turn costs change as dialogs deepen. In addition to dialog-level metrics, LangSmith tracing  can be enabled to capture per-call prompts and responses. The supervisor branches further provide node-level LangGraph traces that record routing decisions and the specialist agents dispatched at each step.

\section{Experimental Results}

\subsection{Evaluation Results}

Table~\ref{tab:evaluation_results} reports the overall evaluation results for the two architectures. 
The supervisor-specialist architecture consistently outperforms the baseline across both subjective and objective metrics, indicating improvements not only in response quality but also in tool-calling reliability. 
The gains in planning effectiveness (+54.5\%) and task completion (+37.8\%) suggest that the supervisory layer helps the agent decompose complex requests, coordinate specialist modules, and synthesize intermediate results more effectively.

The objective metrics further show that these quality improvements are accompanied by more robust execution behavior. 
Compared with the baseline, our system achieves near-perfect tool-name validity and schema compliance, while also improving execution success and recovery success. 
These results suggest that the proposed design reduces both reasoning-level and system-level failure modes, making it better suited for complex multi-step operational tasks.

\begin{table*}[h]
\centering
\caption{Evaluation results comparing different architectures.}
\label{tab:evaluation_results}
\vspace{4pt}
\renewcommand{\arraystretch}{1.15}
\resizebox{\textwidth}{!}{%
\begin{tabular}{lccccccc}
\toprule
\textbf{Architecture} 
& \textbf{Plan. Eff.} 
& \textbf{Tool Qual.} 
& \textbf{Task Comp.} 
& \textbf{Name Val.} 
& \textbf{Schema Comp.} 
& \textbf{Exec. Succ.} 
& \textbf{Recovery SR} \\
\midrule
Plan-Execute 
& 0.538 
& 0.791 
& 0.617 
& 0.975 
& 0.927 
& 0.917 
& 0.900 \\

Supervisor-Specialist 
& 0.831 
& 0.917 
& 0.850 
& 1.000 
& 0.997 
& 0.973 
& 1.000 \\
\bottomrule
\end{tabular}%
}
\end{table*}

\begin{figure}[h]
    \centering
    \includegraphics[width=\columnwidth]{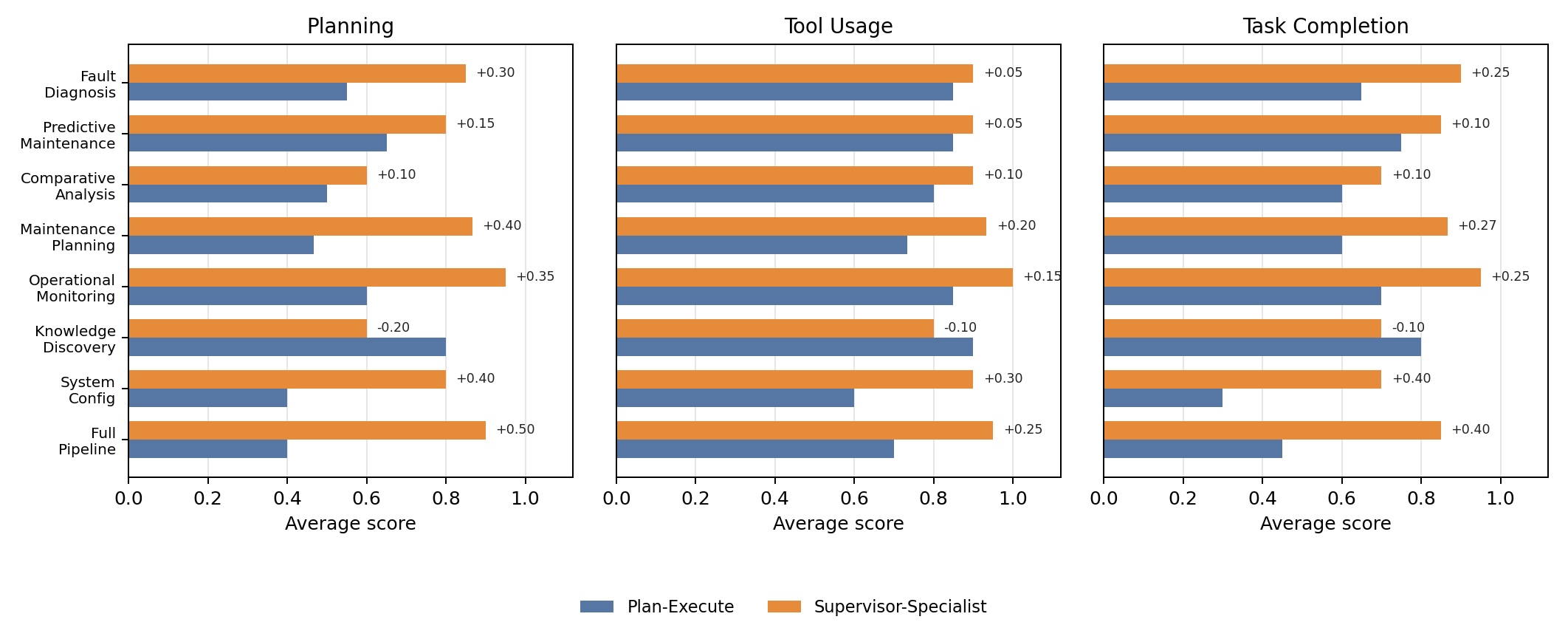}
    \caption{Category-level comparison of subjective evaluation scores for the plan-execute and supervisor-specialist architectures. Scores are averaged over dialogs within each task category for planning effectiveness, tool usage quality, and task completion. Values annotated beside the supervisor-specialist bars indicate the score difference relative to plan-execute.}
    \label{fig:improvement_vs_baseline}
\end{figure}

Beyond the aggregate results, Fig.~\ref{fig:improvement_vs_baseline} reveals clear variation in architectural advantages across task categories. The supervisor-specialist agent achieves higher average scores in most categories, with the largest gains observed in Full Pipeline, System Configuration, Maintenance Planning, and Operational Monitoring tasks. These categories typically require coordinating multiple subtasks, selecting appropriate specialists, and integrating evidence across tool outputs, which aligns well with the supervisor-specialist architecture. In contrast, plan-execute performs better in Knowledge Discovery / Onboarding, where tasks mainly involve a short, structured sequence of information retrieval steps rather than complex cross-agent coordination. This suggests that explicit plan execution remains effective for simple lookup-style workflows, whereas the supervisor-specialist design offers stronger benefits for multi-step diagnostic and operational tasks requiring adaptive decomposition and synthesis. Detailed category-level averages are provided in Appendix~\ref{app:category_average_scores}.

In summary, the supervisor-specialist architecture demonstrates stronger overall effectiveness, reliability, and robustness than the baseline, particularly for operational tasks that require adaptive decomposition and coordination. The results also indicate that its advantages are most pronounced when tasks involve multiple interdependent steps and tool interactions. Overall, these findings support our design as a more suitable architecture for complex, tool-centric agent workflows.

\subsection{Profiling Results}

\subsubsection{Run-Level Summary}

Table~\ref{tab:profiling_run} summarizes the completion statistics and resource consumption of the three architectures across 16 dialogs.

\begin{table}[H]
\centering
\caption{Run-level profiling summary.}
\label{tab:profiling_run}
\renewcommand{\arraystretch}{1.15}
\small
\begin{tabular}{lccc}
\toprule
\textbf{Metric} 
& \textbf{Plan-Execute} 
& \textbf{Supervisor-Specialist} 
& \textbf{Supervisor-Specialist (Parallel)} \\
\midrule
Total wall time (min)  
& 83.9 
& \textbf{65.2} 
& 73.3 \\

Total tokens consumed  
& \textbf{2.55M} 
& 3.32M 
& 3.62M \\

Total LLM API calls    
& 841 
& 941 
& \textbf{751} \\
\bottomrule
\end{tabular}
\end{table}

The plan-execute baseline consumes the fewest tokens, however, it requires the longest total wall time, suggesting that lower token consumption does not necessarily correspond to higher execution efficiency. 
By contrast, the supervisor-specialist architecture achieves the shortest wall time, completing all dialogs in 65.2 minutes. 
This suggests that the architecture improves end-to-end runtime performance by reducing execution errors and the associated overhead from re-planning, retries, and recovery, despite its additional token usage.

The parallel supervisor-specialist variant achieves the lowest number of LLM API calls, reflecting the benefit of concurrent tool execution in reducing sequential interaction overhead. 
Nevertheless, it incurs the highest token consumption and does not achieve the fastest completion time. 
This is likely because parallel tool execution shifts the bottleneck from tool latency to LLM generation, which is further analyzed in Section~4.2.4.
Overall, the profiling results highlight a clear trade-off: the plan-execute baseline is most token-efficient, the standard supervisor-specialist architecture is most time-efficient, and the parallel variant reduces API-call frequency at the cost of greater token usage.

\subsubsection{Wall-Time Latency Decomposition}

\begin{figure}[h]
\centering
\includegraphics[width=\linewidth]{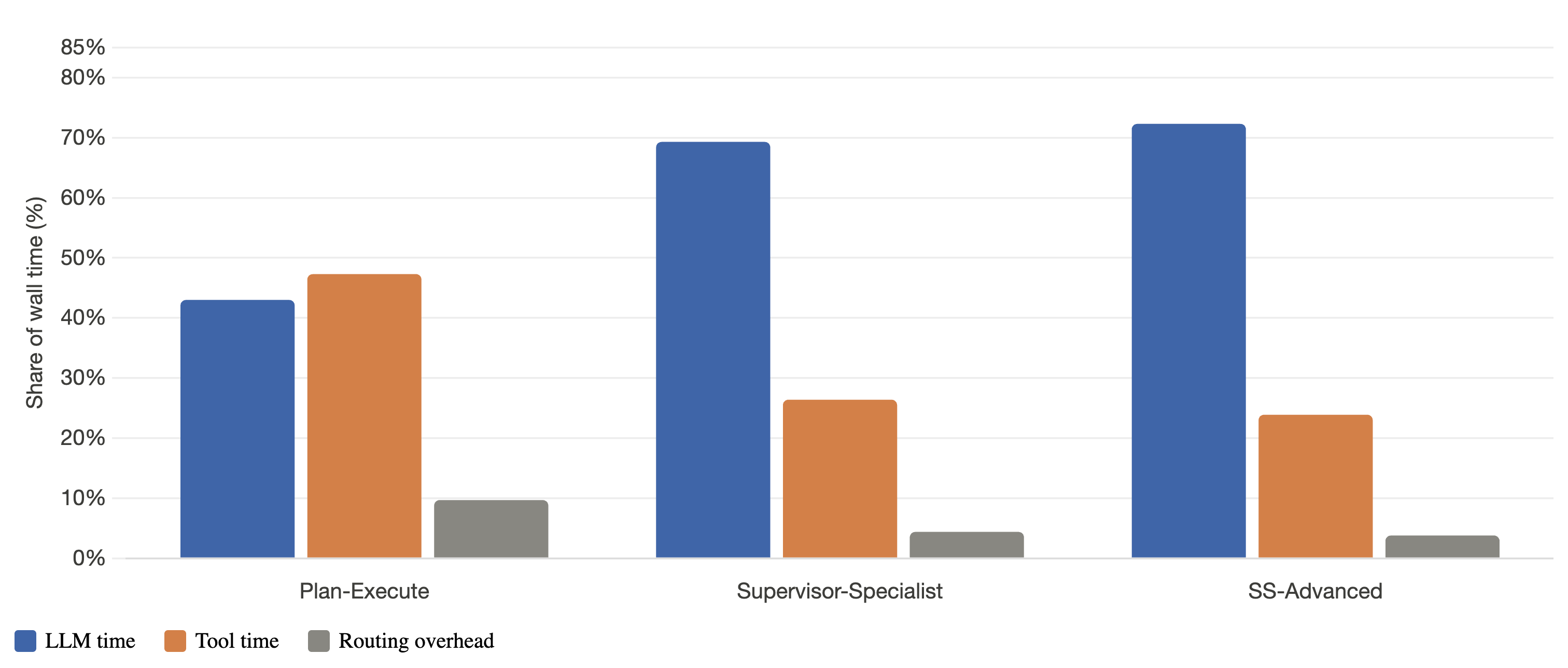}
\caption{Wall-time latency decomposition by architecture. }
\label{fig:latency_breakdown}
\end{figure}

Fig.~\ref{fig:latency_breakdown} reports the three-way latency
breakdown, averaged across all 16 dialogs.

Tool invocation is the main bottleneck in the baseline, accounting for 47.3\% of total wall time, with an average of 186.6\,s per dialog. 
The two supervisor-specialist architectures reduce the tool-time fraction to approximately 25\%. 
Notably, the standard supervisor-specialist architecture and its parallel variant achieve similar tool-time fractions, at 26.4\% and 23.9\%, respectively, despite the use of parallel tool execution in the latter. 
This indicates that supervisor routing, rather than tool-level parallelism, is the main driver of improved tool efficiency. 
As tool overhead is reduced, LLM API latency becomes the dominant bottleneck, contributing 69--72\% of total wall time.

\subsubsection{TSFM as the Critical Tool Bottleneck}

The main architectural difference in tool latency is concentrated in the time-series foundation model (TSFM) server. 
On average, the baseline spends 159.5\,s per dialog on this server, which is more than four times the latency of the standard supervisor-specialist architecture and its parallel variant, at 37.4 and 36.2\,s, respectively. 
Moreover, this server alone accounts for 85.5\% of the baseline's total tool time, with some dialogs lasting as long as 495\,s. 
This suggests that the plan-execute reasoning strategy may issue redundant or insufficiently bounded time-series queries when multi-step reasoning diverges. 
In contrast, the supervisor-specialist architectures route requests to the time-series analysis specialist more precisely and terminate querying once sufficient evidence has been obtained.

Latency on the IoT server also decreases, from 24.9\,s in the baseline to 12.9\,s in the standard supervisor-specialist architecture, mainly due to the removal of duplicated sensor-history queries. 
By comparison, the work-order, utilities, and FMSR servers show little architectural difference in absolute latency. 
Their increased proportional share under the supervisor-specialist architectures is therefore primarily a denominator effect caused by the substantial reduction in time-series foundation model server latency.

\begin{figure}[h]
\centering
\includegraphics[width=\linewidth]{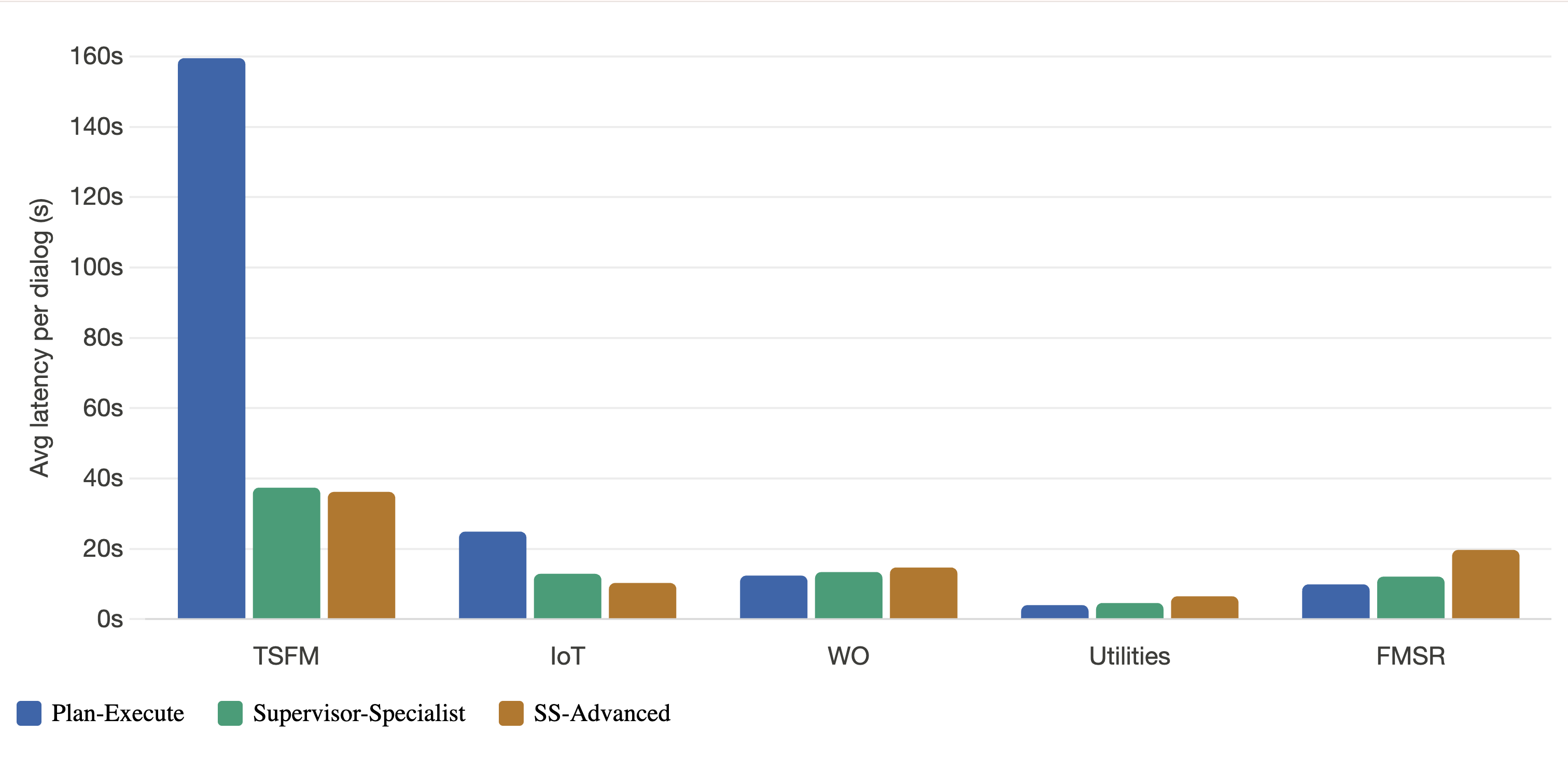}
\caption{Average MCP tool server latency per dialog.}
\label{fig:server_latency}
\end{figure}

\subsubsection{Context Growth and Token Cost}

The parallel supervisor-specialist architecture suffers from context bloat caused by the aggregation of parallel tool outputs. 
After all parallel tool invocations return, their outputs are concatenated to form the input to the next LLM invocation. 
This substantially increases prompt size: the average number of tokens per invocation is 4,825, 59\% higher than the plan-execute baseline at 3,036, while the 95th percentile reaches 14,272 tokens, more than twice the baseline value of 6,361.

The effect is especially severe in some dialogs. 
For instance, dialog~3 under the parallel supervisor-specialist architecture consumes 1,438,115 tokens in total, which is 3.5$\times$ larger than the second-largest case across all architectures. 
This explains why parallel tool execution does not effectively reduce timeout risk. 
As context size grows, each LLM invocation becomes slower, shifting the bottleneck from tool latency to LLM generation. 
Accordingly, the parallel supervisor-specialist architecture shows the longest single invocation time of 122.7\,s, compared with 44.1\,s for the baseline, and a slow-call rate ($>$10\,s) of 6.9\%, compared with 2.0\% for the baseline.

\subsubsection{Per-Turn Cost Distribution}

Table~\ref{tab:profiling_turn} shows average turn durations by position within each dialog.

\begin{table}[H]
\centering
\caption{Average turn duration by position (seconds).}
\label{tab:profiling_turn}
\renewcommand{\arraystretch}{1.15}
\small
\begin{tabular}{lccc}
\toprule
\textbf{Turn} 
& \textbf{Plan-Execute} 
& \textbf{Supervisor-Specialist} 
& \textbf{Supervisor-Specialist (Parallel)} \\
\midrule
1          
& 69.3  
& 145.4 
& 169.7 \\

2          
& 89.6  
& 18.4  
& 14.4  \\

3          
& 90.9  
& 37.5  
& 55.5  \\

4          
& 59.6  
& 26.2  
& 61.8  \\

5          
& 105.7 
& 184.0 
& 26.9  \\

\rowcolor{gray!15}
Turns 2--5 avg.
& 83.4 
& \textbf{34.3} 
& 40.9 \\
\bottomrule
\end{tabular}
\end{table}

The supervisor-specialist architectures front-load execution cost into the first turn because of LangGraph initialization and cold-start routing dispatch. 
The standard supervisor-specialist architecture takes 145\,s in turn~1, while the parallel supervisor-specialist variant takes 170\,s. 
Afterward, their average turn latency decreases to 34.3\,s and 40.9\,s, respectively, as the graph warms up and intermediate artifacts are reused, yielding an approximately 4.2$\times$ speedup.

The baseline shows the opposite trend: its first turn is the cheapest at 69.3\,s, whereas later turns average 83.4\,s. 
Its response length also grows monotonically from 1,100 to 6,700 characters, a 6$\times$ increase, suggesting unbounded context accumulation across turns. 
Both supervisor-specialist architectures maintain stable response lengths of 2,700--3,200 characters throughout the dialogs.

\section{Discussion}

\subsection{Interpretation of Results}
The results show that the supervisor-specialist architecture improves multi-turn industrial O\&M dialog mainly by enabling structured task decomposition and cross-turn artifact reuse. Compared with the plan-execute baseline, our system produces better planning, tool usage, and task completion, while also reducing schema and execution failures. The profiling results further suggest that the key efficiency gain comes from avoiding redundant tool calls, especially repeated time-series forecasting queries. Although the supervisor introduces more overhead in the first turn, this cost is amortized in subsequent turns because previously generated evidence can be reused for follow-up questions.

\subsection{Limitations}

Several limitations remain in the current study. 
First, the evaluation is based on 16 benchmark dialogs and a limited set of chiller assets, so the results may not fully capture the diversity of real industrial sites, failure modes, and user behaviors. 
Second, the system relies on available benchmark data and predefined MCP tool interfaces, which may be cleaner and more structured than those in production O\&M environments. 
Third, the parallel tool execution setting increases prompt size and token consumption, shifting the bottleneck from tool latency to LLM generation latency. 
As a result, the parallel supervisor-specialist architecture does not consistently outperform the standard supervisor-specialist architecture.

\subsection{Future Work}
Future work should focus on improving both system efficiency and evaluation coverage. On the system side, parallelism should move from low-level tool calls to parallel specialist execution, combined with evidence summarization, pruning, and selective artifact retrieval to reduce prompt bloat. On the evaluation side, future benchmarks should include larger and more realistic industrial datasets, longer dialogs, and memory-specific metrics such as artifact reuse rate, memory recall accuracy, redundant tool-call reduction, and cross-turn consistency. These improvements would help determine whether the system can generalize from controlled benchmark scenarios to real industrial O\&M workflows.

\section{Conclusion}

This paper presents a multi-turn dialog system for industrial asset operations and maintenance based on a supervisor-specialist multi-agent architecture. 
The system addresses key limitations of the plan-execute single-agent baseline, including weak cross-turn context management, fragile tool invocation, and redundant retrieval of expensive operational evidence. 
By decomposing diagnostic workflows into specialist agents and storing their outputs as reusable artifacts, the proposed architecture supports memory-aware replanning and more reliable tool use across follow-up questions.

The evaluation results show that the supervisor-specialist architecture improves both response quality and execution reliability. 
Compared with the baseline, it improves planning effectiveness, tool usage quality, and task completion, while also increasing schema compliance and execution success. 
The profiling results further show that the main efficiency gain comes from reducing redundant tool usage, especially repeated calls to the time-series foundation model server. 
Although the supervisor-specialist architecture incurs higher first-turn overhead, later turns become substantially faster because previously generated artifacts can be reused.

The results also show that parallel tool execution is not automatically beneficial. 
In our experiments, the parallel supervisor-specialist architecture reduces some tool-side latency but increases token consumption and LLM generation time because of larger aggregated contexts. 
This finding suggests that future system optimization should focus not only on parallelism, but also on compact evidence representation, selective memory retrieval, and efficient inter-agent communication.

Overall, this work demonstrates that multi-agent decomposition and structured artifact reuse are effective design principles for tool-centric industrial O\&M dialog systems. 
While broader datasets and more realistic deployment conditions are still needed, the proposed approach provides a promising foundation for building robust, efficient, and context-aware assistants for industrial asset operations.

\section*{Acknowledgments}

We are grateful to Dr. Dhaval Patel and Dr. Kaoutar El Maghraoui from IBM Research for their valuable guidance and mentorship throughout this work.

\clearpage

\bibliographystyle{plain}
\bibliography{references}

\clearpage

\appendix

\section{Evaluation Metrics}
\label{app:evaluation_metrics}

\begin{table*}[!htbp]
\centering
\caption{Evaluation metrics for the multi-turn industrial dialog system. Subjective metrics are assessed by LLM-as-Judge, while recovery and objective metrics are computed using rule-based evaluation.}
\label{tab:evaluation_metrics}
\small
\setlength{\tabcolsep}{4pt}
\vspace{4pt}
\renewcommand{\arraystretch}{1.15}
\begin{tabularx}{\textwidth}{
>{\raggedright\arraybackslash}p{0.18\textwidth}
>{\raggedright\arraybackslash}p{0.22\textwidth}
>{\raggedright\arraybackslash}X
}
\toprule
\textbf{Metric} & \textbf{Level} & \textbf{Aggregation and Description} \\
\midrule

Planning Effectiveness 
& Dialog-level subjective metric 
& Macro-averaged across dialogs. Evaluates whether the agent decomposes the task properly, maintains goal alignment, and re-plans when necessary. \\

Tool Usage Quality 
& Dialog-level subjective metric 
& Macro-averaged across dialogs. Evaluates whether the agent selects appropriate tools, constructs valid arguments, and uses context effectively across turns. \\

Task Completion 
& Dialog-level subjective metric 
& Macro-averaged across dialogs. Measures whether the final response successfully addresses the user's diagnosis or maintenance planning needs. \\

Tool Name Validity 
& Call-level objective metric 
& Micro-averaged over all extracted tool calls. Measures the percentage of tool calls whose tool names exist in the available tool set, indicating whether the agent avoids hallucinating unavailable tools. \\

Schema Compliance 
& Call-level objective metric 
& Micro-averaged over valid-name tool calls. Measures the percentage of valid-name tool calls whose arguments satisfy the required schema and structured input constraints. \\

Execution Success 
& Call-level objective metric 
& Micro-averaged over all extracted tool calls. Measures the percentage of tool calls that execute successfully without runtime failure. \\

Recovery Success Rate 
& Dialog-level recovery metric 
& Computed over dialogs where a recovery action was taken. Measures the percentage of such dialogs that eventually complete successfully after retrying, correcting invalid arguments, or re-planning. \\

\bottomrule
\end{tabularx}
\end{table*}

\section{Category Average Table}
\label{app:category_average_scores}

\begin{table*}[!htbp]
\centering
\caption{Category-level average subjective evaluation scores. P, T, and C denote Planning Effectiveness, Tool Usage Quality, and Task Completion, respectively. $\Delta$ denotes Supervisor-Specialist minus Plan-Execute.}
\label{tab:category_average_scores}
\small
\setlength{\tabcolsep}{4pt}
\vspace{4pt}
\renewcommand{\arraystretch}{1.15}
\begin{tabularx}{\textwidth}{
>{\raggedright\arraybackslash}X
c
*{9}{>{\centering\arraybackslash}p{0.68cm}}
}
\toprule
\textbf{Category} 
& \textbf{\#Dialogs}
& \multicolumn{3}{c}{\textbf{Plan-Execute}}
& \multicolumn{3}{c}{\textbf{Supervisor-Specialist}}
& \multicolumn{3}{c}{\textbf{$\Delta$}} \\
\cmidrule(lr){3-5}
\cmidrule(lr){6-8}
\cmidrule(lr){9-11}
& & P & T & C & P & T & C & P & T & C \\
\midrule

Fault Diagnosis 
& 4 & 0.55 & 0.85 & 0.65 & 0.85 & 0.90 & 0.90 & +0.30 & +0.05 & +0.25 \\

Predictive Maintenance 
& 2 & 0.65 & 0.85 & 0.75 & 0.80 & 0.90 & 0.85 & +0.15 & +0.05 & +0.10 \\

Comparative Analysis 
& 1 & 0.50 & 0.80 & 0.60 & 0.60 & 0.90 & 0.70 & +0.10 & +0.10 & +0.10 \\

Maintenance Planning 
& 3 & 0.47 & 0.73 & 0.60 & 0.87 & 0.93 & 0.87 & +0.40 & +0.20 & +0.27 \\

Operational Monitoring 
& 2 & 0.60 & 0.85 & 0.70 & 0.95 & 1.00 & 0.95 & +0.35 & +0.15 & +0.25 \\

Knowledge Discovery / Onboarding 
& 1 & 0.80 & 0.90 & 0.80 & 0.60 & 0.80 & 0.70 & -0.20 & -0.10 & -0.10 \\

System Configuration 
& 1 & 0.40 & 0.60 & 0.30 & 0.80 & 0.90 & 0.70 & +0.40 & +0.30 & +0.40 \\

Full Pipeline (End-to-End) 
& 2 & 0.40 & 0.70 & 0.45 & 0.90 & 0.95 & 0.85 & +0.50 & +0.25 & +0.40 \\

\midrule
\textbf{Overall} 
& \textbf{16} 
& \textbf{0.54} & \textbf{0.79} & \textbf{0.62} 
& \textbf{0.83} & \textbf{0.92} & \textbf{0.85} 
& \textbf{+0.29} & \textbf{+0.13} & \textbf{+0.23} \\
\bottomrule
\end{tabularx}
\end{table*}

\end{document}